\documentclass[preprint,authoryear,12pt]{elsarticle}

\usepackage{epsfig}
\usepackage{epstopdf}

\usepackage{graphics}

\usepackage{graphicx}

\usepackage{amssymb}
\usepackage{graphicx,array,colortbl,color,rotating,amsmath,amssymb,afterpage}
\usepackage{multirow}
\usepackage{epsfig}
\usepackage{enumerate}
\usepackage{hyphenat}
\usepackage{verbatim} 
\usepackage{rotating}
\usepackage{url}
\usepackage{multicol}
\usepackage{algorithm2e}
\usepackage{dsfont}

\newdefinition{example}{Example}
\newdefinition{definition}{Definition}
\newdefinition{property}{Property}

\DeclareMathOperator*{\argmax}{argmax}

\newcommand{\Perf}{\mathit{Perf}}

\journal{}

\begin{document}

\begin{frontmatter}

\title{Simulating Non Stationary Operators in Search Algorithms}

\author{Adrien Go{\"e}ffon}
\address{LERIA, University of Angers (France)} 
\author{Fr{\'e}d{\'e}ric Lardeux}
\address{LERIA, University of Angers (France)}
\author{Fr\'ed\'eric Saubion}
\address{LERIA, University of Angers (France)}

\begin{abstract}
In this paper, we propose a model for simulating search operators whose behaviour often changes continuously during the search. In these scenarios, the performance of the operators decreases when they are applied. This is motivated by the fact that operators for optimization problems are often roughly classified into exploitation operators and exploration operators. Our simulation model is used to compare the different performances of operator selection policies and clearly identify their ability to adapt to such specific operators behaviours. The experimental study provides interesting results on the respective behaviours of operator selection policies when faced to such non stationary search scenarios.

\end{abstract}


\end{frontmatter}

\section{Introduction}

Selecting the most suitable operators in a search algorithm when solving optimization problems is an active research area \citep{Eiben2007,lobo07}. Given an optimization problem, a search algorithm mainly consists in applying basic solving operators --- heuristics ---  in order to explore and exploit the search space for retrieving  solutions. The choice of the successive operators along the search process is often driven by means of parameters. The improvement of the performance of the algorithm thus relies on an adequate setting of these parameters. An optimal setting may be achieved by an optimal  operator selection policy. Unfortunately, according to the underlying intuitions provided by the No Free Lunch theorems for optimization \citep{Wolpert1997a}, this optimal policy may strongly depend on the instances of the problems to be solved. Initial parameters setting can be achieved by automated tuning algorithms \citep{hutter_paramils:_2009,revac-PPSN08}. Nevertheless, the  values of the parameters may require more continuous control \citep{fialho_adaptive_2010} and should rather not be fixed during the whole search process. Adaptive operator selection is strongly related to reinforcement learning problems, and especially to multi-armed bandit problems \citep{Fialho08,Costa2008}. Various methods for managing the famous {\em exploration vs.~exploitation} balance in search heuristics have been investigated in the literature; see for instance \citep{cec2009,lobo07,Thierens2005}. The performance of adaptive selection policies depends on the characteristics of the problem's search space, as well as on the specificities of the search operators. Therefore different families of practical problems have been handled, but also more abstract operators models in order to provide more general and comprehensive testing frameworks as in \cite{Thierens2005} and \cite{Costa2008}, taking into account changes in the operators behaviours. 

\bigskip
\noindent{\bf Motivations}

In this paper, we propose an alternative model for simulating search operators whose behaviour often change continuously during the search. In these scenarios, the performance of the operators decreases when they are applied. This is motivated by the fact that operators for optimization problems are often roughly classified into exploitation operators and exploration operators. Exploitation operators  aim at focusing on the evaluation of the visited configurations of the search space in order to converge quickly to a local optimum. Exploration operators aim at diversifying the search trajectory by visiting sparse areas of the search space. Unfortunately, it is not possible to always exploit nor explore the search space. For instance, it is unlikely that an exploitation operator will always improve a configuration and find directly an optimal solution (except for simple problems). Therefore, decreasing performance may be observed along the search as well as changing behaviours of operators. 

\bigskip
\noindent{\bf Contributions}

Our simulation model is used to compare the different performances of operator selection policies and clearly identify their ability to adapt to such specific operators behaviours  and can be used as a surrogate operator model when designing new adaptive search algorithms. Hence, the general  description  of operator based search algorithms may be helpful in this design process when the user has to precisely identify the components and performance criteria that are used in the adaptive process. The experimental study provides interesting results on the respective behaviours of operator selection policies when faced to such non stationary search scenarios. 
Considered as a multi-armed bandit problem, our model corresponds to a specific restless bandit problem that could be used to model different real applications as soon as the efficiency of a given action decreases according to successive frequent uses. For instance, such reinforcement learning techniques are used to schedule on-line advertisement display on web pages. Our model could be pertinent in this context since it may be clear that the relevance of an advertisement decreases if it is too much shown to the same user. Other cases of such decrease repeated actions may actually be observed in various application domains. 

\bigskip
\noindent{\bf Organization of the paper}

In section \ref{sec:algo}, we describe optimization algorithm that are based on applications of basic search operators. We also define the problem of designing the best possible operator selection policy and show its relationship with multi-armed bandit problems. Section \ref{sec:policies} is dedicated to review different operator selection policies.  Section  \ref{sec:scenarios} presents our model for simulating non stationary operators. Experiments are presented in section \ref{sec:expe}.

\section{Operator Based Search Algorithms for Optimization Problems}

\label{sec:algo}

In this section, we propose to precisely define the components of a search algorithm in the context of solving optimization problems, in order to be able to handle their behaviour. We focus on algorithms that use operators --- also commonly called heuristics --- in order to find solutions in the search space. We introduce the notion of policy, that defines how the operators should be scheduled in a search process, and some criteria in order to compare different policies and thus to characterize an optimal policy. 

\begin{definition}[Optimization Problem]
An optimization problem is a pair $({\cal S},f)$ where ${\cal S}$ is a search space whose elements represent solutions (or configurations) of the problem and $f: {\cal S} \to \mathds{R}$ is an objective function. An optimal solution (for maximization problems) is an element $s^* \in {\cal S}$ such that $\forall s \in {\cal S}, f(s^*) \geqslant f(s).$
\end{definition}

\begin{example}
As a simple running example, we may consider the toy {\em One-Max} problem, where  ${\cal S} = \{0,1\}^n$, for a given size $n$. Given a solution $s =( s_1,\ldots, s_n) \in {\cal S}$, $s_i$ is thus the value of the $i^{th}$ bit of $s$. The objective function is simply $f(s) = \Sigma_{i=1}^n  s_i$. 
\end{example}

\noindent
A general solving algorithm may be  abstracted according to the following components. 

\begin{definition}[Operator Based Algorithm (OBA)]
Given an optimization problem $({\cal S},f)$, a solving algorithm is defined by a tuple $(Init,\Omega,\theta,\pi)$:
\begin{itemize}
\item 
an initialization function $Init : 2^{\cal S} \to 2^{\cal S}$,
\item
a set of operators $\Omega = \{ o_1,\ldots,o_n \}$, which are functions $o_i : 2^{\cal S} \to 2^{\cal S}$,
\item 
a parameter vector  $\theta \in \Theta$, where $\Theta$ is a parameter space\footnote{According to the classification proposed in \citep{smit09comparing}, parameters can be numerical or symbolic.},
\item 
a search policy $\pi : \Theta \times I \to \Omega$. $\pi \in \Pi$, where $\Pi$ is the set of search policies and $I = \{ 1,\ldots,T\} \subseteq \mathds{N}$ is a set of iterations of the algorithm.
\end{itemize}
\end{definition}

Typically, the initialization function takes as input\footnote{This input represents indeed the knowledge that is required by the initialization function.} the whole search space --- or a part  of it --- and returns a random element or a set of random elements of this search space. From a practical point of view, initialization is often insured by a random generation of points of the search space or by a greedy algorithm that builds a suboptimal solution. Note that we do not want to distinguish the different solving paradigms that use different data structures for their search processes, e.g., tree-based search, neighborhood search, population-based search... We only pay attention to the fact that these algorithms use basic search operators (branching heuristics, constraint propagation, hill climbing, recombination...) that they apply on configurations of the search space.  The search policy determines the operator to apply at each iteration of the algorithm. Therefore the notion of iteration is here directly linked to the application of one operator in a steady state fashion; this general model can anyway be adapted for coarser granularities of iterations.  

\begin{example}
Back to  our running example, let us consider a simple evolutionary algorithm for the One-Max problem. The algorithm includes an initialization function such that $Init(S)={\cal P}_0$, where ${\cal P}_0$ is an initial population, randomly generated. The set of operators may include flipping operators (e.g., 1-flip is an operator that flips one bit --- position --- of a configuration $s \in {\cal S}$) and crossover operators (e.g., uniform crossover). Note that here, operators are applied on a population and include a selection and an insertion process. For instance, let ${\cal P} = \{000,111,110\}$, $\hbox{1-flip}({\cal P}) = \{010,111,110\}$. The parameter vector could be $\theta = (psize,\sigma_{1},\ldots,\sigma_{n})$ where $psize$ is the size of the population ${\cal P}$, and $\sigma_{i}$ is the probability of application of an operator $o_i$. The search policy may be $\pi(\theta,t)=o_i$ with a probability $\sigma_{i}$, that corresponds to a roulette wheel selection as described in Section \ref{sec:policies}. Note that, for the moment, we consider that $\theta$ remains constant over iterations. 
\end{example}

The operational semantics of a search algorithm can be defined by means of its runs. We define $\overline{{\cal S}}$ (resp. $\overline{\Omega}$ ) as the set of sequences over $2^{\cal S}$ (resp. $\Omega$). 

\begin{definition}[Run of an OBA]
\label{def:run}
A run of length $l$ of an OBA  $A=(Init,\Omega,$ $\theta,\pi)$ for an optimization problem $P=({\cal S},f)$ is defined as a pair $(\overline{s},\overline{o}) \in \overline{{\cal S}} \times \overline{\Omega}$  where:
\begin{itemize}
\item 
$\overline{s} = s^{(0)},s^{(1)},\ldots,s^{(l)}$ with $s^{(0)}=Init({\cal S})$
\item
$\overline{o} = o^{(1)},\ldots,o^{(l)}$ 
\item 
$\forall i \geqslant 1, o^{(t)}=\pi(\theta,t), \hbox{ with }  $
\item
$\forall i \geqslant 1, s^{(i)}=o^{(i)}(s^{(i-1)})$
\end{itemize}
The set of all possible runs of length $l$ is denoted $Run_l(A,P)$. 
\end{definition}

Given an OBA  $A$ and an optimization problem $P$, the operational semantics of the algorithm is thus the set $Run(A,P)=\bigcup_{l \geqslant 0} Run_l(A,P)$. The runs may have different lengths depending on the stop criteria, e.g. reaching a solution, exceeding a maximum number of operators applications... Of course, if the algorithm is fully deterministic, this set is a singleton (i.e., a single run for a given problem).  

Given a run $r \in Run(A,P)$, we suppose that there exists a performance function such that $\Perf(r) \in \mathds{R}$. A run $r$ is considered better than a run $r'$ if $\Perf(r) \geqslant \Perf(r')$. This performance function is also supposed to be extended to subsets of runs $R \subseteq Run(A,P)$. This performance function may vary according to the type of algorithm and the type of problem (satisfaction, optimisation...). We now define a notion of optimality for the performance of the algorithms that corresponds to classic tuning point of view \citep{Hoos2012}. Various methods have been proposed to this aim \citep{BSP02,smit09comparing,hutter_paramils:_2009}. 

\begin{definition}[Optimal Tuning]
Given an optimization problem $P$, an OBA $A=(Init,\Omega,\theta,\pi)$ is optimally tuned for a problem $P$ iff $\forall \theta' \in  \Theta$, $Perf(Run(Init,\Omega,\theta',\pi),P) \leqslant Perf(Run(Init,\Omega,\theta,\pi),P)$. 
\end{definition}

Note that we decide here to consider tuning only on parameters in $\theta$ and not on the possible policies, which could also be consider as structural parameters --- components --- of the algorithm. 

Only algorithms with fixed parameters and policy have been considered in the previous definitions. Of course, we have to take into account adaptive algorithms in which parameters may change during the search. Note that, in order to simplify our presentation, we do not consider changing policies. As defined above, a policy uses parameters and is then intrinsically submitted to adaptive changes by means of these parameters. Therefore, we may generalize our definition of an OBA to an adaptive OBA,. 

\begin{definition}[Adaptive OBA]
An adaptive operator based algorithm is a tuple  $A=(Init,\Omega,\theta,\pi,K)$ where $(Init,\Omega,\theta,\pi)$ is an OBA and $K$ is a control function $K:\Theta \times I \to \Theta$ that provides a new parameter vector with regards to the current iteration.  
\end{definition}

The notion of run and operational semantics defined above can be extended straightforwardly to adaptive OBAs. We may now turn to optimal control of adaptive OBAs. 

\begin{definition}[Optimal Control]
Given an optimization problem $P=({\cal S},f)$, an adaptive OBA $(I,\Omega,\theta,\pi,K)$ is optimally controlled for a problem $P$ iff $\forall K'$, $\Perf(Run(Init,\Omega,\theta,\pi,K'),P) \leqslant \Perf(Run(Init,\Omega,\theta,\pi,K),P)$. 
\end{definition}

\subsection{Abstracting Operators}

We focus now on the control of the algorithms through their search policies and control functions. Therefore, we consider the notion of adaptive control policy as a pair $(\pi,K)$, which associates a search policy and a parameter control function.  Following the terminology used in the literature \citep{lobo07}, such a policy is called an adaptive operator selection (AOS) policy. Note that the case where $K$ is the identity function corresponds to a non adaptive policy.  Therefore, in the remaining of the paper, we consider OS policies, including adaptive policies and fixed policies cases. 

\begin{example}
Back to the example, we may consider an adaptive roulette wheel instead of the fixed one as previously mentioned. The control function may thus update the probability according to the observed performances of the operators by means of the function $K$, which is known as probability matching (see Section \ref{sec:polprob}). 
\end{example}

Given an optimization problem $P$, our purpose is to generate a policy $(\pi,K)$ that produces optimal runs $(\overline{s},\overline{o})$, which may be abstracted as optimal sequences of operator applications $\overline{o}$. It is clear that the impact of the operators depends on the elements of  $\overline{s}$ on which they are applied. We consider that each operator provides a gain when it is applied; this gain may  be evaluated according to the performance function.

\begin{definition}[Gain of an operator]
 \label{def:gaiN_{op}p}

We define a gain function $g : \Omega \times \overline{{\cal S}} \to \mathds{R}$, for an operator $o$, as follows.

$$g(o,\overline{s})=\Perf((\overline{s}\oplus o(s^{(k)}),\overline{o}\oplus o)) -\Perf((\overline{s},\overline{o}))$$

where $\overline{s}=s^{(0)},\ldots,s^{(k)}$; $\overline{o}=o^{(1)},\ldots,o^{(k)}$; $\overline{s}\oplus o(s^{(k)}) = s^{(0)},\ldots,s^{(k)},o(s^{(k)})$ and $\overline{o}\oplus o = o^{(1)},\ldots,o^{(k)}, o$. 
\end{definition}

\noindent
Note that the gain of an operator is defined according to the variation of gain that it provides on the current run. The gain of a policy can be computed for one run (definition \ref{def:gainPolicy}).

\begin{definition}[Gain of an OS Policy]
\label{def:gainPolicy}
Given a problem $P$ and an OBA $ (Init,\Omega,\theta,\pi,K)$, the gain obtained by the OS policy is thus the sum 

$$G((\pi,K),r)= \sum_{o^{(k)} \in \overline{o}} g(o^{(k)},(s^{(0)},\ldots,s^{(k)}))$$

 such that $r=(\overline{s},\overline{o}) \in Run((Init,\Omega,\theta,\pi,K),P)$.    
\end{definition}

\noindent
The operator selection policy problem can be defined as follows. 

\begin{definition}[OS Policy Problem]
Given $P=({\cal S},f)$, the components $Init$, $\Omega$ and $\theta$ of an OBA,  the operator selection policy problem consists in finding a policy 

$(\pi^*,K^*) = \underset{\pi \in \Pi,K \in {\cal K}}{\argmax}   ~\sum_{r \in Run(A,P)} G((\pi,K),r)$ 

with $A=(Init,\Omega,\theta,\pi,K)$
\end{definition}

Of course since most of the time,  the whole set $Run(A,P)$ cannot be extensively computed, the optimal policy has to be approximated on a subset of $Run(A,P)$. From a practical point of view, there is a connection between the notion of optimal control and the notion of optimal policy. 

Searching for an optimal policy which chooses at each step operators that maximize the overall gain may be related to bandit problems \citep{Costa2008,fialho_adaptive_2010}.  

\subsection{ Multi-Armed Bandits}
\label{sec:MAB}

The initial stochastic multi-armed bandit (MAB) \citep{Robbins1952,Bradt1956,Rodman1978} is formulated as follow. Given several possible actions --- usually  called arms according to the gambling machine analogy --- that have different individual gains (or rewards), one has to select a sequence of actions  that maximizes the total gain. Definition \ref{def:MAB} proposes a more formal definition of this general problem.

\begin{definition}[Stochastic MAB]
 \label{def:MAB}
Let us consider $n$ independent arms. For each arm $i \in \{1,\ldots,n \}$, we have: 
\begin{itemize}
\item
a set of possible states $S_i$;
\item
a set of probabilities $Prob_i = \{ \sigma^i_{j\rightarrow k}, j,k \in S_i \}$ such that  $\sigma^i_{j\rightarrow k}$ is the probability of being in state $k$ if the arm $i$ is played\footnote{We use the verb {\em play} according {\em gain} to the gambling analogy.} from state $j$;
\item
a set of gains $G_i =\{g^i_j, j \in S_i\}$ where $g^i_j$ is the gain obtained when arm $i$ is played from state $j$.
\end{itemize}
\end{definition}

Given a stochastic MAB, the problem is to find a policy that maximizes over a finite\footnote{Note that we restrict the problem to finite horizon MAB. The most general problem is often presented over infinite horizon.} horizon $T$, $\Sigma_{t=0}^{T} g_t\gamma^t$, where $g_t$ is the expected gain of the policy at time $t$ and $\gamma \in [0,1]$ is a discount factor. 

Four features can be identified to characterize a MAB problem  \citep{Mahajan2008}: 
\begin{enumerate}
\item only one arm is played at each time; 
\item states of unplayed arms do not change;
\item arms are independent;
\item arms that are not played do not contribute any gain.
\end{enumerate}

Many variants of the initial stochastic MAB have been studied in the literature. In this paper we focus on the restless MAB, first introduced in \citep{Whittle1988}. In this formulation, the gains of the arms change over time, while they are supposed to be fixed --- but of course unknown --- in the initial stochastic MAB formulation. In fact, restless bandits may be defined as in Definition \ref{def:MAB} except that, when an arm is not played, its state may change, which corresponds to a relaxation of Feature 2. Hence, restless MABs involve two kinds of probabilities in $Prob_i$, namely $\sigma^i_{j\rightarrow k}$, which represents the probability of being in state $k$ if the arm $i$ is played, and $\tilde{\sigma}^i_{j\rightarrow k}$, that is the probability of being in state $k$ if the arm $i$ is not played.

\section{Operator Selection Policies}
\label{sec:policies}

In this sections we first explain how operator selection in operator based algorithms  can be directly related to the choice of the most suitable sequence of actions in the context of multi-armed bandit problems. We review then different possible selection policies that can be used to achieved an optimal schedule of the operators.  

\subsection{Operator Selection in Operator Based Algorithms}
\label{sec:aos}

Let us consider an OBA  $A=(I,\Omega,\theta,\pi,K)$. Here, we are not interested in the initialization function $I$. Let $\Omega = \{o_1, \ldots, o_n\}$ be the set of $n$ operators. We have to define the control policy $(\pi,K)$ which selects an operator at each iteration of the algorithm in order to build a run $(\overline{s},\overline{o})$. We review here different policies and we distinguish between policies based on probabilities of application of the operators and policies based on upper confidence bounds.

The gain of an operator (see Definition \ref{def:gaiN_{op}p}) is generally specific to the problem, since it uses the notion of performance of a run. In order to have a more general approach, a general notion of utility\footnote{Note that we use the term utility here, which should be clearly related to the notion of action value in reinforcement learning \citep{Sutton1998}.}, which reflects the successive gains obtained by the operators, can be introduced. 

Considering a run $(\overline{s},\overline{o})$, such that $\overline{s}=s^{(0)},\ldots,s^{(n)}$ and $\overline{o}=o^{(1)},\ldots,o^{(n)}$, an utility $u_{i}^{(t)}$ is associated to each operator $i \in \{1..n\}$ for any iteration $t \in \{1..n\}$.  This utility has to be re-evaluated at each time, classically using a formula $u_{i}^{(t)} = (1-\alpha) u_{i}^{(t-1)} +\alpha.g(o_i,s^{(0)},\ldots,s^{(t-1)})$, with $u_{i}^{(0)} = 0$. This utility uses the gain associated to the application of operator $i$ (which corresponds thus to the immediate utility) and $\alpha$ which is a coefficient that controls the balance between past and immediate utilities, as in classic reinforcement learning techniques \citep{Sutton1998}. If an operator is not selected at iteration $t$, its gain is $0$ for this iteration. 

\subsubsection{Policies based on probabilities of application}
\label{sec:polprob}

In this context, given the set of operators $\Omega = \{o_1, \ldots, o_n\}$, we use the parameter vector $\theta$ to associate a probability of selecting the operator, $\theta = (\sigma_1,\ldots,\sigma_n)$ such that $\Sigma_{i=1}^n \sigma_i = 1$. The search policy $\pi$ is then a roulette selection wheel that selects each operator $o_i$ according to its probability of selection $\sigma_i$. Different operator selection policies  have been proposed in the literature \citep{lobo07,Hamadi2012}; we review here some of the most used of them. 
\begin{itemize}
\item
{\bf Fixed Roulette Wheel}

A first possibility consists in keeping $\theta$ fixed during the run, i.e. $\forall t \in I, K(\theta,t)=\theta$. Note that these values can be determined by an automated tuning process \citep{Hoos2012}.  
\item
{\bf Adaptive Roulette Wheel}

 Contrary to a static tuning of the operator application rates, adaptive operator selection consists in selecting the next operator to apply at iteration $t+1$ by adapting the selection probability during the search. In this case, we have $\theta^{(t)} = (\sigma_{1}^{(t)},\ldots,\sigma_{n}^{(t)})$. The control function $K: \Theta \times I \to \Theta$ is defined as  $K(\theta^{(t)},t+1)=\theta^{(t+1)}$. Defining $K$ consists in defining the probabilities $\sigma_i ^{(t+1)}$ with regards to the evolution of the operator's utilities. 

A classic mechanism is the probability matching selection rule: 

$\sigma_i ^{(t+1)} = p_{\min} + (1-n.p_{\min}) \frac{u_{i}^{(t+1)}}{\Sigma_{k=1}^{n} u_{i}^{(t+1)}}$

where a non negative value $p_{\min}$ insures a non zero selection probability for all operators. Note that, in order to insure a coherent behaviour,  $p_{\min}$ should be in the interval $[0,\frac{1}{n}]$. 
\item
{\bf Adaptive Pursuit}

An  alternative proportional selection rule has been proposed in \cite{Thierens2005}, called adaptive pursuit (AP), that distinguishes the best current operator from the others: 

\[
\left\{
\begin{array}{l}
\sigma_{i^*}^{(t+1)} = \sigma_{i^*}^{(t)} + \beta (p_{\max} - \sigma_{i^*}^{(t)}) \\
\sigma_{i}^{(t+1)} = \sigma_{i}^{(t)} + \beta (p_{\min} - \sigma_{i}^{(t)}) \\
\end{array}
\right.
\]
where $i^* \in \underset{i \in \{1,\ldots,n \}} {\argmax} ~u_{i}^{(t+1)}$, $p_{\max} = 1-(n-1)p_{\min}$ and $\beta$ is a parameter to adjust balance of this winner-take-all strategy. 
\end{itemize}

\subsubsection{Policies based on upper confidence bounds}
\label{sec:polucb}

Optimal strategies have been initially proposed  by \cite{Feldman1962} and \cite{Gittins1979} for the multi-armed bandit problem. Later, \cite{Auer2002b} proposed to use this problem to manage the compromise between exploration and exploitation in optimization algorithms. The following policies consists in computing an upper confidence bound of the expected gain and to select thus the most promising arm. 

\begin{itemize}
\item
{\bf UCB (upper confidence bound)}

The UCB1 criterion \citep{Auer2002} is defined as: 

$\forall o_i \in \Omega, UCB1(o_i,t) = u_{i}^{(t)} + \sqrt{\frac{2 \log (\sum_{1 \leqslant k \leqslant n} nb_{k}^{(t)})}{nb_{i}^{(t)}}} $

where $nb_{i}^{(t)}$ denotes the number of times operator $o_i$ has been applied. Note that this formula is defined for gains that should be normalized between $0$ and $1$. The left term of the formula uses the successive utilities that are obtained by the arms in order to focus of the best arm, while the right term aims at providing the opportunity to be selected for less used arms. This formula attempts thus to achieve a compromise between exploitation and exploration.  

Therefore, we may define the control policy $(\pi,K)$ for a given iteration $t$ as: 

\[
\left\{
\begin{array}{l}
\pi(\theta^{(t)},t) =  \underset{i \in \{1,\ldots,n\}}{\argmax} ~\theta_i^{(t)}\\
K(\theta^{(t)},t) = \theta^{(t+1)} = (UCB1(o_1,t+1),\ldots,UCB1(o_n,t+1))\\
\end{array}
\right.
\]

Here no parameter is required. Note that UCB has originally be designed for fixed gain distributions. Since the gain of the operators is likely to change along the search, UCB has been extended to dynamic multi-armed bandit has to be considered.  

\item
{\bf DMAB (Dynamic MAB algorithm based on UCB)}

UCB has been revisited in \cite{Costa2008}. A standard test --- known as  Page Hinkley \citep{Hinkley1970} --- for the change hypothesis is used.  We may add a parameter in $\theta$ which indicates if the process has to be restarted. In this case the  control function $K$ use the Page Hinkley test to detect statistical changes in the successive utilities of the operators and may re-initialize the values of the operators utilities.  Moreover, it can be useful to add a scaling factor to the right term of the UCB1 formula in order to take into account the value range for utilities. The test is parametrized by $\gamma$ that controls its sensitivity and $\delta$ that manages its robustness. We refer the reader to \cite{Fialho2010} for more details.

\end{itemize}

\subsection{Policies based on Transition Matrix}
\label{sec:polim}

Based on previous work on island models for operators selection \citep{gecco12}, we present here a selection policy based on a probabilistic transition matrix. The underlying motivation is not only to detect the best possible operators but also possible relationships between operators. 

Given an OBA $A$, we define a matrix $M$ of size $|\Omega|\times |\Omega|$.  $M(o_i,o_j)$ represents the probability of applying operator $o_j$ after having applied $o_i$. According to our previous notations, $M$ corresponds to the $\theta$ parameter of the selection policy (introducing a two-dimensional parameter structure). 

From an operational point of view, contrary to previous policies, this policy uses simultaneously several possible runs of the OBA in order to acquire its knowledge. Let us consider a set $P^{(t)}$ of $psize$ runs of length $t+1$ produced by the algorithm $A$ at iteration $t$ (remind that runs are numbered from index $0$ in definition \ref{def:run}). $psize$ is a parameter of this policy. For each operator $o_i \in \Omega$, let $P_i^{(t)}$ be the set of runs $(s^{(0)}\ldots s^{(t)},o^{(1)}\ldots o^{(t)}) \in P^{(t)}$ where $o^{(t)}=o_i$. Hence, this set is the set of all runs whose last applied operator is $o_i$, and we have  $P^{(t)} = \underset{o \in \Omega}{\bigcup} P_o^{(t)}$.

The control policy $(\pi,K)$ can thus be defined from $M$. Firstly, the policy $\pi$ is defined by a roulette selection whose probabilities are given by $M$. Secondly, the update of $M$ is performed by the control function $K$ as: 

\[
\begin{array}{l c l}
K(M^{(t)}(i,k),t)& =& M^{(t+1)}(i,k)\\
& =& (1-\beta)(\alpha.M^{(t)}(i,k) + (1-\alpha)R^{(t)}_{i}(k)) + \beta N^{(t)}(k)
\end{array}
\]
where $N^{(t)}$ is a stochastic vector such that $||N^{(t)}||=1$ and $R^{(t)}_{i}$ is the reward vector that is computed by using the utility of the operators that have been used after applying $i$. More precisely: 

$$
 R^{(t)}_{i}(k) =
  \begin{cases}
  \frac{1}{|B|} & \text{if $k \in B$},\\
  0 & \text{otherwise},
  \end{cases}
$$ $$
\text{with\ } B = \underset{o_k \in \Omega}{\argmax} ~\bigg (\max_{\{ \overline{s} \in \overline{{\cal S}}| \exists r \in P_k^{(t)}, r = (\overline{s},  o^{(1)} \ldots o^{(t-1)}=o_i, o^{(t)}= o_k) \}}   g(o_k,\overline{s}) \bigg )
$$

$B$ is the set of the best operators that have been applied after an operator $o_i$, i.e. that have provided the best gain. We could also update $R^{(t)}_{i}(k)$ by using the mean of the improvements. Remark also that this policy does not use the utility which is indeed computed within the update process. 

The parameter $\alpha$ represents the importance of the knowledge accumulated (inertia or exploitation) and $\beta$ is the amount of noise, which is necessary to explore alternative possibilities. The influence of these parameters has been studied in \cite{gecco12}.

 This approach can be related to reinforcement techniques for MDP (Markov Decision Processes) \citep{Sutton1998}. Nevertheless island models use several populations in order to learn simultaneously from several sequences of operators.

\section{Modelling Scenarios for Gain Functions}
\label{sec:scenarios}

In practice,  the gain function $g$ may be difficult to compute and is very specific to the problem at hand. Therefore  we may approximate such a gain function  by using distributions. Comparisons of operator selection policies would be then easier and faster.  As previously defined, we use a gain function $g(o,t)$ which represents an estimation of the gain of an operator $o$ if it is applied at iteration $t$.  Several scenarios for modelling $g$ can  be envisioned. 

In Definition  \ref{def:gaiN_{op}p}, the gain associated to an operator is defined according to the performance improvements that  it provides during a given run of the algorithm. Remind that, for an optimization problem  $P=(f,{\cal S})$, the effect of operators can be defined according to the evaluation function $f$.  For instance, given an element $s \in {\cal S}$, the variation of evaluation $f$ induced by an operator $o$ is given as  $f(o(s))-f(s)$.  This function can be extended to $\overline{{\cal S}}$. Of course different performance criteria could be taken into account.

\subsection{Fixed  Gains}

For each operator $o$, the gain $g$ is defined independently from iteration $t$ by a binomial probability distribution $(p_o,g_o)$, i.e., $\forall i \in I, \Pr[g(o,i) = g_o] = p_0$ and $\Pr[g(o,i) = 0] = (1-p_o)$. Such distributions have been proposed in \cite{Costa2008} for studying operator selection policies. 

In this context, if the values $(p_o,g_o)$ are fixed during the whole run of the algorithm, then determining the best policy just consists in finding the operator that has the greatest expected gain $p_o . g_o$. This is indeed a basic stochastic multi-armed bandit problem (Definition \ref{def:MAB}), where operators correspond to arms with only one state. Of course, it is unlikely that, in real optimization problems, the effect of an operator remains unaltered in a whole run. Therefore it would be more realistic and interesting to consider gain functions that may evolve during the run.

\subsection{Epoch Based Gains} 

In \cite{Thierens2005}, uniform distributions are associated to each operator in different overlapping intervals of values. The distributions are fixed during a given number of iterations --- called epoch --- and then are re-assigned according to a permutation. The gain of the operator is thus non stationary during the  algorithm's execution and the AOS has to discover the best operator to apply at each epoch. Such scenarios have been studied using various techniques including adaptive pursuit \citep{thierens07operatorAllocation}, dynamic UCB \citep{Fialho08} and genetic algorithms \citep{Koulouriotis2008}. 

\subsection{Non stationary gains with sliding time windows} 

In this section we define a new scenario in order to model operators whose behaviour changes more continuously during the solving process. We want to consider more continuous changes in the gain distributions. The idea is to provide a model where the gain of an operator decreases proportionally to its use. In such a model, the AOS policy must not detect the best operator during an epoch but rather identify suitable sequences of operators. 

Within a run $(\overline{s},\overline{o})$ of length $n$ of an OBA, such that $\overline{o} = o_1, \ldots,o_n$, we compute the gain of operator $o$ at iteration $t$ thanks to a gain function $g_{wsize}(o,t)$, defined as:
$$g_{wsize}(o,t)  = g_o .\bigg(1-\frac{Occ_{wsize}(\overline{s},o,t)}{wsize}\bigg)$$

$g_o$ is a fixed maximal gain of operator $o$, and $Occ_{wsize}(\overline{s},o,t)$ is the number of applications of operator $o$ during the last $wsize$ iterations. More formally, $Occ_{wsize}(\overline{s},o,t) = |\{i \in 1..|\overline{w}|, \omega_i=o\}|\} $, where $\overline{w}$ is the subsequence $o_{t-wsize},\ldots, o_{t-1}$ that records the $wsize$ last applied operators. $wsize$ is a fixed  parameter of the scenario. 
 
\noindent
This scenario  can be formalized as a restless MAB problem: 

\begin{itemize}
\item
An arm $i$ is associated to each operator $o_i \in \Omega$.
\item 
For each operator $o_i \in \Omega$, we define a set of states $S_i = \{0..(2^{( wsize+1)}-1)\}$, such that each state represents the previous use of the arm by a binary number. For instance, for a window of size $wsize = 4$, the state $1001$ means that the arm has been played at iteration $t-1$ and $t-4$ only. 
\item 
The set of transition probabilities $Prob_i$ between states can be defined as: 
\begin{enumerate}
\item
$
\sigma^i_{j\rightarrow k} =
\begin{cases}
1 & \mathrm{if}\ k = Lshift(j)+1\\
0 & \mathrm{otherwise}\\
\end{cases}
$

\item
$
\tilde{\sigma}^i_{j\rightarrow k} =
\begin{cases}
1 & \mathrm{if}\ k = Lshift(j)\\
0 & \mathrm{otherwise}\\
\end{cases}
$
\end{enumerate}

$Lshift(j) = $ denotes the logical left shift of a binary number of fixed size $wsize$. Note that we need $\tilde{\sigma}$ probabilities since we are modelling a restless MAB problem (see definition \ref{def:MAB} and section \ref{sec:MAB}).  
\item 
When an arm $i$ is played from state $j$, the reward can be  straightforwardly defined  as $r^i_j = g_o.(1-\frac{\#_{1}(j)}{wsize})$, where 
 $\#_{1}(j)$ is the number of bits being equal to $1$ in state $j$. 
\end{itemize}

This restless bandit problem involves a two states transition matrix whose size is in $O((n.2^{wsize})^2)$. Of course in practice, one has to memorize the $wsize$ previous applications of operators.

\subsection{ Non Stationary Binary Scenarios}
\label{sec:nsbs}

In optimization algorithms, operators are often classified  intuitively in two classes : diversification operators and intensification operators, whose behaviours should be understood as orthogonal. In order to better understand such scenarios we consider here two types of operators, according to the previous notations: $(1,1)$  and $(1,0)$, that we denote here respectively $\mathds{1}$ and $\mathds{O}$. The $\mathds{1}$ operators theoretically always gain 1 but their efficiency will decrease proportionally to their use according to our sliding window. $\mathds{O}$ always gain $0$. We choose here probability of $1$ for all operators in order to avoid probability side effects in our analysis. We also use gain values whose range is in $[0,1]$.

An instance of a non stationary binary scenario can be fully  defined by a triple  $(N_{op},N_1,wsize)$ where $N_{op}$ is the number of operators, $N_1$ is the number of $\mathds{1}$ operators and $wsize$ is the length of the window used for computing the decay of the operator's gains (see section \ref{sec:policies}). 
. 

It can be shown that, independently from the size of the windows, the score obtained by a uniform choice of the operators is constant: 

\begin{property}
Given an instance $(N_{op},N_1,wsize)$, the expectation of gain for a $\mathds{1}$ operator is:

$$\forall t \in I, \mathbb{E}[g(\mathds{1},t)] = 1-\frac{1}{N_{op}}$$
\end{property}

\noindent
Indeed, $\Pr[Occ_{wsize}(\overline{s},o,t) = k] = \big(\frac{1}{N_{op}})^k  \big(1-\frac{1}{N_{op}} \big)^{wsize-k} \binom{wsize}{k}$, with a corresponding gain of  $\big(1 - \frac{k}{wsize}\big).p.r$ for a $(p,r)$ operator, i.e. $\big(1 - \frac{k}{wsize}\big)$ for a $\mathds{1}$ operator. Considering that $Occ_{wsize}(\overline{s},o,t)$ is a binomially distributed random variable of parameters $wsize$ and $\frac{1}{N_{op}}$, then $\mathbb{E}[Occ_{wsize}(\overline{s},o,t)] = \frac{wsize}{N_{op}}$, and $\mathbb{E}[g(\mathds{1},t)] = \mathbb{E}\big[1-\frac{Occ_{wsize}(\overline{s},o,t)}{wsize}\big] = 1-\frac{1}{N_{op}}$. 

\noindent
More generally, we have then the following property:
\begin{property}
Given an instance $J\equiv (N_{op},N_1,wsize)$, the gain expectation per iteration is equal to $\frac{N_1}{N_{op}}\big(1-\frac{1}{N_{op}}\big)$. 
\end{property}

Therefore the behaviour of a uniform selection of the operators is independent from $wsize$, for a given $N_{op}$. The total expected gain for $N_1$ operators can easily be computed according to the number of allowed iterations. A uniform choice policy may  thus serve as baseline for other selection processes (see Section \ref{sec:expe}). 

We may suppose that the $\mathds{1}$ operators are numbered $1.. N_1$ and $\mathds{O}$ are numbered $N_1+1 .. N_{op}$.  The optimal policy  problem can be formulated as a discrete constraint optimization problem. 

\begin{definition}[Optimal Policy for Binary Non Stationary Scenarios]
Given a non stationary binary scenario $(N_{op},N_1,wsize)$ and an horizon $T$ (the number of allowed iterations) we define 

\begin{itemize}
\item
the set of variables ${\cal X} = \{x_1,\ldots, x_T\}$, such that $x_i$ is the operator applied at iteration $i$ (decision variables)
\item 
$\forall i \in \{1, \ldots,T \}, x_i \in \{1,\ldots,N_{op}  \}$ (domains of the decision variables)
\item 
$\forall j \in \{1, \ldots, N_1  \}, g_j= \frac {1}{wsize} \sum_{k=1}^{T-wsize} \bigg ( \sum_{i \in k .. k+wsize | x_i=j} i \bigg )$ (gains for all operators whose value is 1)
\item 
the objective function is $\max ~\Sigma_{j \in \{1.. N_1  \}} g_j$
\end{itemize}
\end{definition}

Due to the size of the induced search space, this problem cannot be solved pratically for large values of $T$. Nevertheless, we may restrict this problem to a limited window and compute suboptimal policies. We consider a restricted model. Note that, since we are only interested in maximizing the gain, the $\mathds{O}$ operators are all equivalent and we may consider only one $\mathds{O}$ operator, i.e., $N_{op} = N_1 +1$ or $N_{op} = N_1$ if no $\mathds{O}$ operator is considered. We consider here a circular scenario of length $Sc$ on which the total gain is computed in order to simulate the successive repetitions of this scenario. Since we want to be able to compute circularly the total gain induced by such a scenario, we have to consider a computation sequence $comps$ that is both a multiple of the length of the windows $wsize$ and of the length of the scenario $Sc$ (i.e., the least common multiple $lcm$).  

\begin{definition}[Suboptimal Policy for Binary NS Scenarios]
\label{def:subopt}

Given a non stationary binary scenario $(N_{op},N_1,wsize)$ and scenario length $Sc$  (the number of allowed iterations of operators in the scenario) we define 

\begin{itemize}
\item 
$comps = lcm(wsize,Sc)$ (the length of the computation sequence)
\item
$nbseq = comps \hbox{ div } Sc$  (the number of sequence of length $Sc$ in the total computation windows)
\item
the set of variables ${\cal X} = \{x_1,\ldots, x_{comps}\}$, such that $x_i$ is the operator applied at iteration $i$ (decision variables)
\item 
$\forall i \in \{ 1, \ldots,comps \}, x_i \in \{1, \ldots, N_{op}  \}$ (domains of the decision variables)

\item
$\forall i \in \{ 1, \ldots , comp \}, k \in \{ 1, \ldots, nbseq-1\},  (x_i=x_{i+(k*Sc)})$ (the scenario is repeated $nbseq$ times in the computation sequence)
\item
$\forall i \in \{ (wsize+1), \ldots,comps \}$  such that $ x_i \in \{1, \ldots,.. N_1\}$,\\
$g_i =\frac{1}{wsize}\sum_{\{k \in (i-wsize)..i | x_k=x_i \}} k$ (standard gains for operators in the computation sequence)
\item
$\forall i \in \{ 1,\ldots , wsize \}$ such that $ x_i \in \{1, \ldots,.. N_1\}$,\\
$g_i = \frac{1}{wsize} \bigg (\sum_{\{k \in 1..i | x_k=x_i \}} k + \sum_{\{k \in (comps-wsize+i)..comps | x_k=x_i \}} k \bigg )$ (circular case)
\item 
the objective function is $\max ~\sum_{j \in \{1.. N_1  \}} g_j$
\end{itemize}
\end{definition}

This model can be used to compute sub-optimal policies. We have used the Minizinc \citep{minizinc} constraint modelling and solving framework in order to compute such policies. Table \ref{exact1} shows the expected gain per iteration for different scenario window sizes, considering one $\mathds{O}$ operator and one $\mathds{1}$ operator. 

\renewcommand{\tabcolsep}{0.07cm}

\begin{table}
\begin{center}
\begin{footnotesize}
\begin{tabular}{|c|r|r|r|r|r|r|r|r|r|r|r|r|r|r|r|}
\hline
	&\multicolumn{1}{c|}{2}	&\multicolumn{1}{c|}{3}	&\multicolumn{1}{c|}{4}	&\multicolumn{1}{c|}{5}	&\multicolumn{1}{c|}{6}	&\multicolumn{1}{c|}{7}	&\multicolumn{1}{c|}{8}	&\multicolumn{1}{c|}{9}	&\multicolumn{1}{c|}{10}	&\multicolumn{1}{c|}{11}	&\multicolumn{1}{c|}{12}	&\multicolumn{1}{c|}{13}&\multicolumn{1}{c|}{14}	&\multicolumn{1}{c|}{15}\\
\hline
1	&{\bf 0.500}	&0.333	&	{\bf 0.500}&	0.400&	{\bf 0.500}& 0.428 	&	{\bf 0.500}&	0.333&	{\bf 0.500}&	0.454 &	{\bf 0.500}&	0.461&	{\bf 0.500}&0.400\\
2	&0.250	&0.333	&{\bf 0.375}	&0.300	&0.333	&0.357	&{\bf 0.375}	&	0.333 & 0.300	&0.363	&{\bf 0.375}	& 0.346	& 0.357	& 0.333\\
3	&{\bf 0.333}	&0.222	&{\bf 0.333}	&{\bf0.333}	&{\bf 0.333}	&0.286	&{\bf 0.333}	&{\bf 0.333}	&{\bf 0.333} 	& {\bf 0.333}	& {\bf 0.333}	& {\bf 0.333}	& {\bf 0.333}	& {\bf 0.333}\\
4	&0.250	&0.250	&0.250	&0.300	&{\bf 0.333}	&0.321	&0.313	&0.278	&0.300	&0.318	&{\bf 0.333}	& 0.326	&	0.321 & 0.300\\
5	&0.300	&0.267	&0.300	&0.240	&0.300	&0.314	&{\bf 0.325}	&0.311	&0.300	&0.273	&0.300	&0.308	&0.314	&0.320\\
6	&0.250	&0.222	&0.292	&0.267	&0.250	&0.286	&0.313	&0.315	&{\bf 0.317}	&0.303	&0.292	&0.269	&0.286	&0.300\\
7	&0.286	&0.238	&0.286	&0.257	&0.286	&0.035	&0.286	&0.302	&{\bf 0.314}	&0.312	&0.310	&0.297	&0.286	&0.267\\
8	&0.125	&0.250	&0.063	&0.275	&0.135	&0.268	&0.031	&0.278	&0.150	&0.307	&0.078	&{\bf 0.308}	&0.152	&0.292\\
\hline
\end{tabular}
\caption{Expected gain per iteration w.r.t. window size ($wsize$) on lines and computation sequences  ($comps$) on column for $N_{op} = 2$ and $N_1 = 1$}
\label{exact1}
\end{footnotesize}
\end{center}
\end{table}

\section{Experimental Results}
\label{sec:expe}

In this section, we study the behaviour of different operator selection policies on binary non stationary scenarios. Note that, as mentioned above, the behaviour of these policies for fixed gain operators or epoch based scenarios has already been studied and will not be considered here. 

\subsection{Experimental Settings}

The following notations are used for the different policies : 

\begin{itemize}
\item
$GR$ is a basic greedy selection policy that always  selects the operator with the current maximal utility. This policy is used without any learning stage, which cannot be efficient here since the gains of the operators continuously change. 
\item 
$EGR$ is an $\epsilon$-greedy policy that selects greedily the best operator according to its current utility,but  uses an exploration coefficient $\epsilon$. Therefore at each iteration the operator with the current maximal utility is selected with probability $1-\epsilon$ and a randomly selected operator is used with probability $\epsilon$.  Note that $GR$ and $EGR$ are basic reinforcement learning strategies. Despite their poor results, these policies just serve here as baseline to highlight that a simple operator policy cannot be efficient for our scenarios. 
\item
$OR$ is a {\em myopic} oracle that is aware of  the gains of the operators and the last applied operators in order to select the next operator with the best expected value according to the variation of total gain .  Note that this oracle is different from the suboptimal policy that can be computed using Definition \ref{def:subopt}. 

This oracle does not compute the optimal solution for a given number of iterations since it does not take into account future operators applications that are used to compute gains. In particular, compare to the optimal and suboptimal policies described in section \ref{sec:nsbs}, $OR$ does not fully make use of the size of the windows $wsize$. 
\item
$U$ is a uniform selection choice rule.It correspond to a fixed roulette with equal probabilities. 
\item
$ARW$ is an adaptive probability matching selection rule (adaptive roulette wheel) using a minimal probability $p_{\min}$, as defined in Section \ref{sec:polprob}. 
\item
$AP$ is the adaptive pursuit method with parameters $p_{\min}$ and $\beta$ (See section \ref{sec:polprob}).
\item
$UCB$ uses the upper confidence bound UCB1 as described in Section \ref{sec:polucb}.
\item
$DMAB$ is the dynamic selection based on UCB1 with the Page Hinkley test. This method requires several parameters for the test (see description in Section  \ref{sec:polucb}). 
\item
$IM$ is the island model implementation of the policy based on transition  matrix as described in Section \ref{sec:polim}. We do not detail the algorithmic implementation here and refer the reader to \cite{gecco12} for more details. 

\end {itemize}

We first set  the experimental conditions: 

\begin{itemize}
\item
Each policy is run 20 times on each instance. Since island models use $80$ individuals, the mean value (MeanV) corresponds to the average score of the $20$ best individuals obtained at each run. In order to achieve fair comparisons, other policies have been run $20 \times 80$ times and the $20$ best scores have been extracted from each sequence of $80$ runs.  Note that the value $80$ has been chosen here since we use a scenario with $8$ operators and we thus use sub-populations of size $10$ for each operators.

\item
Parameters

\renewcommand{\tabcolsep}{0.35cm}
\begin{center}
\begin{tabular}{|l |l |l|l|}
\hline
Method & Parameters & Range Value & Tuned values\\
\hline
GR & - & - & - \\
\hline
EGR & $\epsilon$ & $[0,1]$ & 0.05\\
\hline
OR & - & - & -\\
\hline
U & - & - & -\\
\hline 
ARW &  $p_{\min}$ & $ [0,1]$ & $0.05$\\
\hline
\multirow{3}{*}{IM}& $\alpha$ & $ [0,1]$ & $0.8$\\
    & $\beta$ & $ [0,1]$ & $0.01$\\
    & $nbind$ & $\mathds{N}$ & $80$\\
\hline
\multirow{2}{*}{AP} & $\beta$ & $ [0,1]$ & $0.7$\\
    & $p_{\min}$ & $ [0,1]$ & $0.1$\\
\hline
UCB & - & - & -\\
\hline
\multirow{2}{*}{DMAB} & $\gamma$ & $[0,\infty]$ & $0$ \\
          &      $\delta$ & $[0,\infty]$ & $0$ \\
\hline
\end{tabular}
\end{center}

\item 
The policies have all been implemented using Scilab \citep{scilab}. Experiments have been run on a desktop computer with Intel Core i5 CPU, $2.6$ GHz, 4 Go RAM. Parameters have been tuned using a principled approach inspired by F-Race \citep{BSP02}. Note that here there are indeed few parameters for these policies. 
\end{itemize}

\subsection{Results}

The following tables \ref{res1} and \ref{res2} compare different methods for windows sizes $wsize$ varying from 1 to 8, using  binary non stationary scenarios with $N_{op} = 8$ and $N_1$ varying from 1 to 8, as defined in Section \ref{sec:nsbs}. We report average scores of the best $20$ runs and their standard deviation.  

\renewcommand{\tabcolsep}{0.17cm}
\begin{table}
\begin{center}
\rotatebox{90}{
\begin{tiny}
\begin{tabular}{|l|r|r|r|r|r|r|r|r|r|r|r|r|r|r|r|r|}
\hline
\multicolumn{17}{|c|}{$wsize$ = 1}\\
\hline
$N_1$	&\multicolumn{2}{c|}{1}	&\multicolumn{2}{c|}{2}	&\multicolumn{2}{c|}{3}	&\multicolumn{2}{c|}{4}	&\multicolumn{2}{c|}{5}	&\multicolumn{2}{c|}{6}	&\multicolumn{2}{c|}{7}	&\multicolumn{2}{c|}{8}\\
\hline
OR 	&476.10	&1.66	&1000.00	&0.00	&1000.00	&0.00	&1000.00	&0.00	&1000.00	&0.00	&1000.00	&0.00	&1000.00	&0.00	&1000.00	&0.00\\
IM	&465.40	&1.96	&955.40	&6.72	&966.60	&3.37	&973.90	&2.08	&982.60	&0.97	&988.80	&1.03	&994.30	&1.34	&998.50	&0.53\\
GR	&1.00	&0.00	&1.00	&0.00	&1.00	&0.00	&1.00	&0.00	&1.00	&0.00	&1.00	&0.00	&1.00	&0.00	&1.00	&0.00\\
EGR	&57.30	&2.00	&227.70	&54.86	&245.90	&40.95	&278.60	&32.64	&318.00	&34.66	&340.20	&28.90	&355.60	&24.11	&365.80	&26.68\\
U 	&129.80	&3.52	&245.60	&4.93	&356.90	&3.38	&472.10	&4.84	&584.30	&8.15	&691.90	&5.26	&794.10	&3.41	&901.80	&4.83\\
UCB	&86.00	&0.00	&526.70	&129.23	&909.10	&96.23	&958.30	&0.67	&969.90	&0.32	&977.00	&0.00	&984.00	&0.00	&991.00	&0.00\\
ARW	&249.30	&2.41	&481.20	&7.54	&594.10	&9.30	&665.30	&3.68	&736.00	&4.40	&793.50	&6.92	&844.70	&4.76	&895.60	&2.99\\
AP	&230.50	&4.84	&330.00	&3.65	&426.20	&2.20	&516.90	&3.81	&611.30	&6.15	&700.70	&5.36	&785.90	&7.31	&876.90	&5.24\\
MAB	&86.00	&0.00	&502.40	&83.03	&922.10	&49.93	&958.70	&0.48	&969.90	&0.32	&977.00	&0.00	&984.00	&0.00	&991.00	&0.00\\
\hline
\hline
\multicolumn{17}{|c|}{$wsize$ =2}\\
\hline
$N_1$	&\multicolumn{2}{c|}{1}	&\multicolumn{2}{c|}{2}	&\multicolumn{2}{c|}{3}	&\multicolumn{2}{c|}{4}	&\multicolumn{2}{c|}{5}	&\multicolumn{2}{c|}{6}	&\multicolumn{2}{c|}{7}	&\multicolumn{2}{c|}{8}\\
\hline
OR 	&324.15	&0.91	&678.10	&1.54	&1000.00	&0.00	&1000.00	&0.00	&1000.00	&0.00	&1000.00	&0.00	&1000.00	&0.00	&1000.00	&0.00\\
IM	&91.95	&3.94	&507.70	&2.34	&956.65	&4.06	&966.35	&3.56	&973.40	&3.50	&979.10	&2.31	&983.05	&1.19	&987.15	&1.27\\
GR	&1.50	&0.00	&1.50	&0.00	&1.50	&0.00	&1.50	&0.00	&1.50	&0.00	&1.50	&0.00	&1.50	&0.00	&1.50	&0.00\\
EGR	&57.30	&3.23	&303.20	&59.41	&342.70	&62.10	&365.90	&26.24	&398.60	&38.25	&448.10	&60.37	&464.75	&47.86	&472.45	&39.13\\
U 	&129.95	&2.23	&244.15	&4.81	&360.50	&5.31	&468.95	&4.42	&580.00	&6.84	&684.50	&4.55	&788.35	&3.27	&892.75	&2.60\\
UCB	&116.00	&0.00	&676.45	&0.37	&700.85	&0.85	&944.10	&21.13	&971.65	&0.24	&979.25	&0.26	&986.50	&0.00	&994.00	&0.00\\
ARW	&245.45	&3.37	&473.95	&2.25	&586.50	&2.40	&661.70	&5.14	&730.05	&4.07	&785.75	&6.17	&838.20	&4.09	&888.45	&2.59\\
AP	&228.35	&4.08	&331.30	&4.53	&429.05	&4.96	&520.45	&7.51	&612.10	&4.46	&704.55	&3.32	&787.40	&4.40	&874.15	&2.90\\
MAB	&116.00	&0.00	&676.40	&0.21	&702.30	&1.46	&948.45	&6.96	&971.70	&0.48	&979.00	&0.00	&986.50	&0.00	&994.00	&0.00\\
\hline
\hline
\multicolumn{17}{|c|}{$wsize$ =3}\\
\hline
$N_1$	&\multicolumn{2}{c|}{1}	&\multicolumn{2}{c|}{2}	&\multicolumn{2}{c|}{3}	&\multicolumn{2}{c|}{4}	&\multicolumn{2}{c|}{5}	&\multicolumn{2}{c|}{6}	&\multicolumn{2}{c|}{7}	&\multicolumn{2}{c|}{8}\\
\hline
OR 	&245.77	&0.35	&667.33	&0.00	&841.07	&0.84	&1000.00	&0.00	&1000.00	&0.00	&1000.00	&0.00	&1000.00	&0.00	&1000.00	&0.00\\
IM	&80.83	&2.46	&643.63	&1.61	&673.50	&2.71	&960.90	&4.27	&966.73	&2.81	&972.53	&3.18	&975.73	&3.01	&978.60	&2.30\\
GR	&2.00	&0.00	&2.00	&0.00	&2.00	&0.00	&2.00	&0.00	&2.00	&0.00	&2.00	&0.00	&2.00	&0.00	&2.00	&0.00\\
EGR	&59.63	&2.25	&314.03	&49.56	&408.07	&50.57	&457.30	&53.18	&505.13	&44.43	&518.27	&26.36	&546.20	&66.45	&549.07	&24.64\\
U 	&127.53	&2.75	&244.47	&3.86	&357.10	&5.05	&467.33	&5.70	&578.27	&3.14	&683.10	&3.12	&790.27	&3.67	&890.70	&2.77\\
UCB	&142.00	&0.00	&592.30	&0.11	&773.30	&7.30	&800.90	&0.72	&965.00	&4.06	&978.90	&0.47	&987.80	&0.23	&995.23	&0.16\\
ARW	&241.13	&1.69	&471.43	&3.94	&581.83	&3.56	&660.60	&6.13	&727.73	&4.55	&783.00	&4.35	&838.90	&3.82	&886.57	&2.85\\
AP	&226.53	&2.71	&332.80	&3.94	&427.90	&5.10	&522.23	&7.28	&614.30	&4.73	&700.80	&4.21	&792.27	&5.65	&875.07	&2.62\\
MAB	&142.00	&0.00	&592.27	&0.14	&770.03	&10.92	&800.60	&1.03	&964.67	&3.23	&978.50	&0.36	&987.60	&0.21	&995.13	&0.17\\
\hline
\hline
\multicolumn{17}{|c|}{$wsize$ =4}	\\
\hline
$N_1$	&\multicolumn{2}{c|}{1}	&\multicolumn{2}{c|}{2}	&\multicolumn{2}{c|}{3}	&\multicolumn{2}{c|}{4}	&\multicolumn{2}{c|}{5}	&\multicolumn{2}{c|}{6}	&\multicolumn{2}{c|}{7}	&\multicolumn{2}{c|}{8}\\
\hline
OR 	&198.53	&0.40	&608.00	&1.58	&807.13	&0.93	&905.95	&0.81	&1000.00	&0.00	&1000.00	&0.00	&1000.00	&0.00	&1000.00	&0.00\\
IM	&85.17	&3.08	&509.30	&1.30	&735.02	&1.52	&762.58	&1.62	&965.50	&3.59	&969.95	&2.83	&973.15	&1.94	&976.92	&2.29\\
GR	&2.50	&0.00	&2.50	&0.00	&2.50	&0.00	&2.50	&0.00	&2.50	&0.00	&2.50	&0.00	&2.50	&0.00	&2.50	&0.00\\
EGR	&58.88	&3.30	&414.82	&47.13	&450.02	&38.76	&486.55	&30.67	&517.70	&25.94	&533.13	&37.05	&570.33	&46.05	&589.15	&42.27\\
U 	&128.97	&3.40	&243.75	&3.76	&356.65	&3.73	&468.70	&6.32	&580.48	&4.35	&682.55	&5.40	&788.05	&3.53	&887.00	&1.70\\
UCB	&164.50	&0.00	&566.00	&1.70	&805.83	&0.65	&835.58	&0.17	&854.70	&1.39	&974.70	&2.21	&987.00	&0.26	&996.17	&0.17\\
ARW	&241.32	&2.49	&471.05	&3.49	&581.15	&4.74	&659.90	&2.20	&727.42	&4.47	&780.85	&3.21	&836.15	&2.04	&885.52	&3.15\\
AP	&225.35	&1.69	&333.38	&4.67	&428.32	&4.01	&522.75	&4.28	&614.13	&3.41	&700.67	&4.47	&792.88	&2.04	&875.10	&2.23\\
MAB	&164.50	&0.00	&565.60	&1.55	&806.20	&0.42	&835.60	&0.17	&854.42	&1.12	&976.25	&0.93	&987.00	&0.26	&996.17	&0.12\\
\hline
\end{tabular}
\end{tiny}
}
\end{center}
\caption{Results for $wsize \in \{1\ldots 4 \}$}
\label{res1}
\end{table}

\begin{table}
\begin{center}
\rotatebox{90}{
\begin{tiny}
\begin{tabular}{|l|r|r|r|r|r|r|r|r|r|r|r|r|r|r|r|r|}
\hline
\multicolumn{17}{|c|}{$wsize$ = 5}\\
\hline
$N_1$	&\multicolumn{2}{c|}{1}	&\multicolumn{2}{c|}{2}	&\multicolumn{2}{c|}{3}	&\multicolumn{2}{c|}{4}	&\multicolumn{2}{c|}{5}	&\multicolumn{2}{c|}{6}	&\multicolumn{2}{c|}{7}	&\multicolumn{2}{c|}{8}\\
\hline
OR 	&166.70	&0.19	&601.20	&0.00	&800.60	&0.00	&872.56	&0.75	&937.56	&0.61	&1000.00	&0.00	&1000.00	&0.00	&1000.00	&0.00\\
IM	&91.52	&4.66	&582.40	&1.23	&768.56	&2.62	&794.54	&2.41	&819.16	&2.70	&970.48	&2.63	&971.58	&3.07	&975.16	&2.57\\
GR	&3.00	&0.00	&3.00	&0.00	&3.00	&0.00	&3.00	&0.00	&3.00	&0.00	&3.00	&0.00	&3.00	&0.00	&3.00	&0.00\\
EGR	&58.20	&2.15	&430.34	&73.93	&484.98	&51.39	&536.28	&19.68	&528.46	&23.46	&572.24	&31.44	&599.74	&10.36	&611.76	&21.17\\
U 	&130.52	&2.92	&243.38	&3.67	&358.28	&6.14	&466.40	&7.12	&576.94	&4.82	&683.56	&7.01	&788.42	& 3.33 	&887.84	&1.35\\
UCB	&184.80	&0.00	&574.94	&1.93	&738.44	&0.86	&854.08	&0.94	&872.42	&0.24	&887.38	&0.50	&974.80	&7.01	&994.22	&0.37\\
ARW	&240.68	&1.75	&469.22	&3.33	&580.20	&3.40	&659.10	&4.04	&724.72	&2.62	&781.58	&3.47	&834.36	&3.61	&883.46	&2.07\\
AP	&226.30	&2.69	&334.38	&4.45	&430.52	&4.35	&526.74	&6.74	&616.26	&7.43	&707.02	&5.61	&794.02	&3.51	&875.70	&1.91\\
MAB	&184.80	&0.00	&574.26	&0.30	&738.52	&0.81	&854.38	&0.81	&872.44	&0.26	&887.20	&0.45	&980.04	&4.16	&994.14	&0.28\\
\hline
\hline
\multicolumn{17}{|c|}{$wsize$ = 6}\\
\hline
$N_1$	&\multicolumn{2}{c|}{1}	&\multicolumn{2}{c|}{2}	&\multicolumn{2}{c|}{3}	&\multicolumn{2}{c|}{4}	&\multicolumn{2}{c|}{5}	&\multicolumn{2}{c|}{6}	&\multicolumn{2}{c|}{7}	&\multicolumn{2}{c|}{8}\\
\hline
OR 	&144.12	&0.27	&578.72	&1.33	&768.27	&0.65	&862.33	&0.66	&909.57	&0.74	&955.85	&0.47	&1000.00	&0.00	&1000.00	&0.00\\
IM	&96.57	&3.11	&510.23	&0.49	&671.08	&0.99	&813.60	&2.15	&835.00	&1.58	&857.20	&3.18	&973.15	&3.86	&974.35	&2.52\\
GR	&3.50	&0.00	&3.50	&0.00	&3.50	&0.00	&3.50	&0.00	&3.50	&0.00	&3.50	&0.00	&3.50	&0.00	&3.50	&0.00\\
EGR	&59.37	&2.32	&433.13	&28.65	&498.98	&34.03	&510.15	&43.94	&568.27	&53.50	&604.88	&23.08	&647.55	&29.97	&662.83	&26.99\\
U 	&128.28	&3.79	&244.98	&4.94	&357.35	&4.60	&467.82	&4.30	&575.32	&4.74	&684.33	&4.22	&787.58	&3.63	&885.80	&1.38\\
UCB	&201.17	&0.00	&561.00	&0.00	&760.25	&0.35	&840.15	&7.30	&880.75	&0.84	&898.58	&0.23	&910.23	&0.27	&974.58	&8.64\\
ARW	&240.33	&2.13	&468.58	&2.82	&579.22	&1.84	&659.07	&3.39	&722.50	&2.74	&781.13	&3.66	&834.12	&3.63	&883.23	&1.69\\
AP	&224.98	&2.37	&335.73	&5.57	&435.22	&6.26	&527.13	&5.30	&615.90	&5.86	&708.10	&4.19	&796.27	&2.89	&877.12	&2.90\\
MAB	&201.17	&0.00	&561.00	&0.00	&759.60	&1.41	&837.42	&6.16	&881.77	&2.46	&898.58	&0.21	&910.10	&0.34	&975.27	&14.07\\
\hline
\hline
\multicolumn{17}{|c|}{$wsize$ = 7}\\
\hline
$N_1$	&\multicolumn{2}{c|}{1}	&\multicolumn{2}{c|}{2}	&\multicolumn{2}{c|}{3}	&\multicolumn{2}{c|}{4}	&\multicolumn{2}{c|}{5}	&\multicolumn{2}{c|}{6}	&\multicolumn{2}{c|}{7}	&\multicolumn{2}{c|}{8}\\
\hline
OR 	&127.13	&0.16	&573.14	&0.00	&756.39	&0.98	&857.71	&0.00	&897.54	&0.70	&932.19	&0.28	&967.11	&0.38	&1000.00	&0.00\\
IM	&100.46	&3.29	&556.69	&1.81	&702.44	&1.34	&825.40	&4.26	&847.14	&1.61	&865.43	&1.70	&886.01	&1.50	&976.46	&3.01\\
GR	&4.00	&0.00	&4.00	&0.00	&4.00	&0.00	&4.00	&0.00	&4.00	&0.00	&4.00	&0.00	&4.00	&0.00	&4.00	&0.00\\
EGR	&59.20	&2.05	&470.93	&34.59	&519.37	&30.66	&542.99	&22.51	&607.64	&38.34	&645.67	&31.45	&675.43	&27.44	&691.09	&18.53\\
U 	&129.63	&2.80	&245.46	&5.64	&355.79	&5.33	&464.36	&4.79	&577.07	&8.15	&682.26	&4.34	&785.00	&3.19	&884.43	&1.14\\
UCB	&217.14	&0.00	&548.91	&0.78	&711.27	&1.30	&806.94	&0.14	&863.39	&0.07	&905.13	&0.85	&917.91	&0.12	&929.51	&0.19\\
ARW	&240.41	&2.14	&467.09	&2.75	&582.86	&6.93	&656.40	&3.26	&721.04	&2.26	&781.59	&4.08	&833.70	&1.99	&882.81	&1.36\\
AP	&224.63	&1.59	&335.39	&3.13	&434.30	&5.25	&527.60	&4.27	&620.79	&3.49	&707.79	&3.87	&795.66	&5.08	&877.26	&2.09\\
MAB	&217.14	&0.00	&548.80	&0.69	&710.41	&1.12	&806.60	&0.57	&863.49	&0.30	&905.33	&1.22	&917.93	&0.29	&929.46	&0.11\\
\hline
\hline
\multicolumn{17}{|c|}{$wsize$ = 8}\\
\hline
$N_1$	&\multicolumn{2}{c|}{1}	&\multicolumn{2}{c|}{2}	&\multicolumn{2}{c|}{3}	&\multicolumn{2}{c|}{4}	&\multicolumn{2}{c|}{5}	&\multicolumn{2}{c|}{6}	&\multicolumn{2}{c|}{7}	&\multicolumn{2}{c|}{8}\\
\hline
OR 	&113.85	&0.13	&563.27	&1.10	&751.13	&0.00	&837.64	&0.61	&893.13	&0.99	&920.81	&0.47	&947.70	&0.59	&974.55	&0.37\\
IM	&100.39	&4.08	&511.00	&1.02	&723.50	&2.75	&759.10	&1.79	&853.79	&2.95	&871.41	&1.30	&888.58	&1.72	&905.36	&1.06\\
GR	&4.50	&0.00	&4.50	&0.00	&4.50	&0.00	&4.50	&0.00	&4.50	&0.00	&4.50	&0.00	&4.50	&0.00	&4.50	&0.00\\
EGR	&60.30	&2.98	&480.61	&21.00	&527.46	&32.21	&592.42	&21.79	&622.77	&31.81	&663.14	&37.08	&690.80	&28.90	&708.16	&11.92\\
U 	&129.57	&3.12	&242.61	&4.62	&356.14	&3.26	&468.79	&7.13	&579.90	&4.52	&681.92	&3.25	&786.23	&4.28	&884.59	&1.02\\
UCB	&229.13	&0.00	&539.89	&0.48	&734.38	&3.73	&790.52	&1.06	&848.60	&0.13	&878.41	&3.08	&923.50	&0.75	&934.95	&0.26\\
ARW	&240.40	&2.61	&468.75	&2.44	&579.44	&2.06	&659.15	&2.24	&726.02	&4.13	&780.99	&2.70	&833.80	&1.85	&882.02	&1.63\\
AP	&224.30	&1.48	&337.06	&4.74	&433.56	&3.32	&529.66	&2.60	&621.54	&2.91	&707.58	&4.07	&797.25	&3.75	&876.99	&1.37\\
MAB	&229.13	&0.00	&539.80	&0.44	&733.01	&4.73	&789.67	&0.43	&848.65	&0.21	&878.75	&3.28	&923.66	&0.51	&934.94	&0.21\\
\hline
\end{tabular}
\end{tiny}
}
\end{center}
\caption{Results for $wsize \in \{5\ldots 8 \}$}
\label{res2}
\end{table}

\begin{figure}
\begin{center}
\begin{tabular}{c}
\begin{minipage}[t]{0.7\textwidth}
\includegraphics[width=\textwidth]{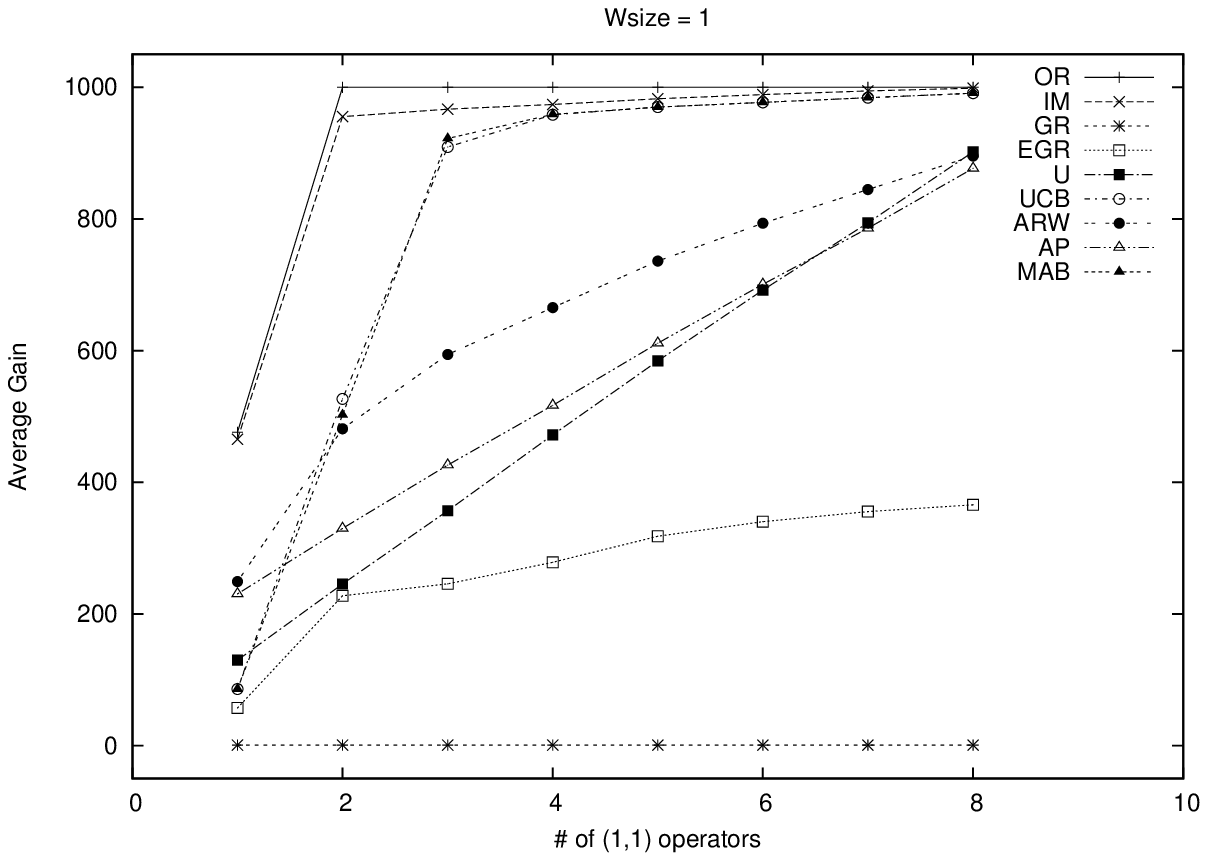}
\end{minipage}
\\
\begin{minipage}[t]{0.7\textwidth}
\includegraphics[width=\textwidth]{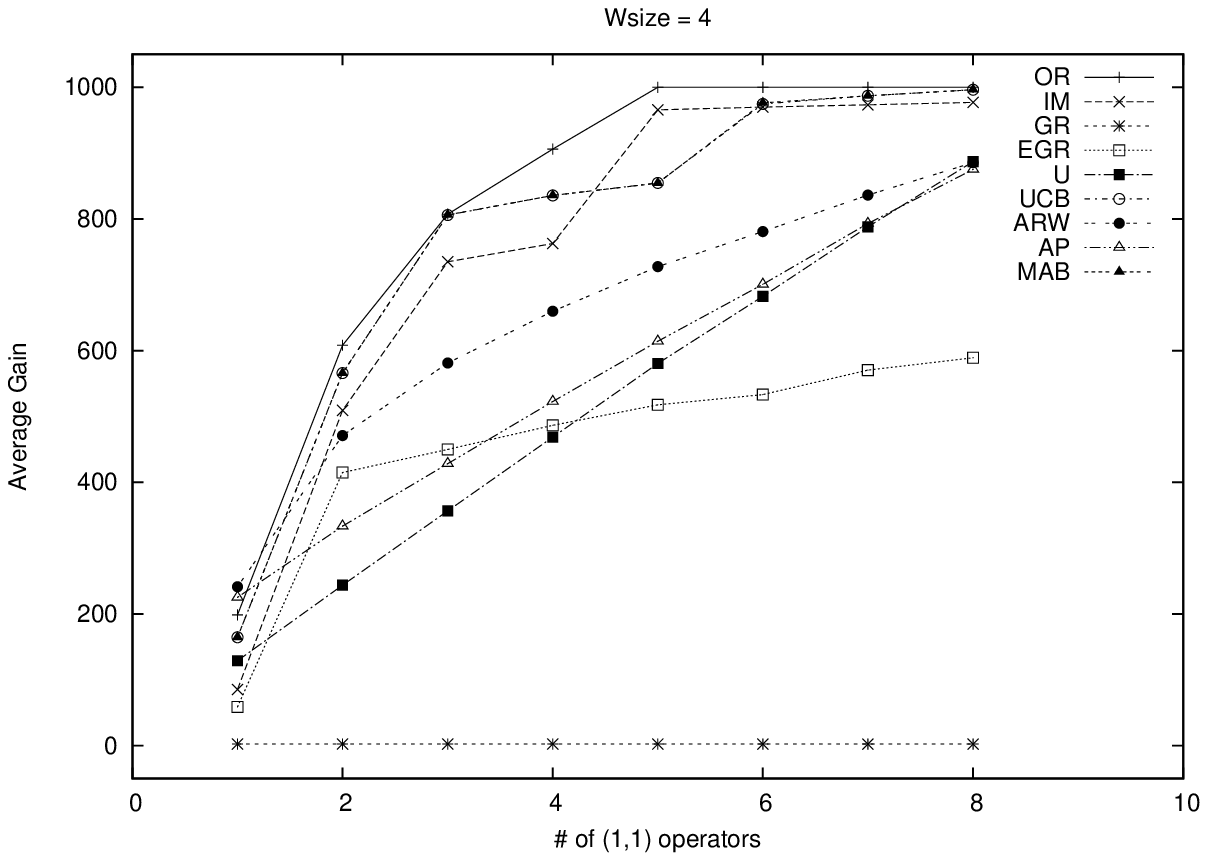}
\end{minipage}
\end{tabular}
\end{center}
\caption{Experiments with different windows size}
\label{fig:results}
\end{figure}

In order to highlight the respective advantages and drawbacks of the different selection policies, Figure \ref{fig:results} propose a graphical view of some of these results.

\subsection{Comments}

\begin{figure}[htb]
\begin{center}
\includegraphics[width=0.7\textwidth]{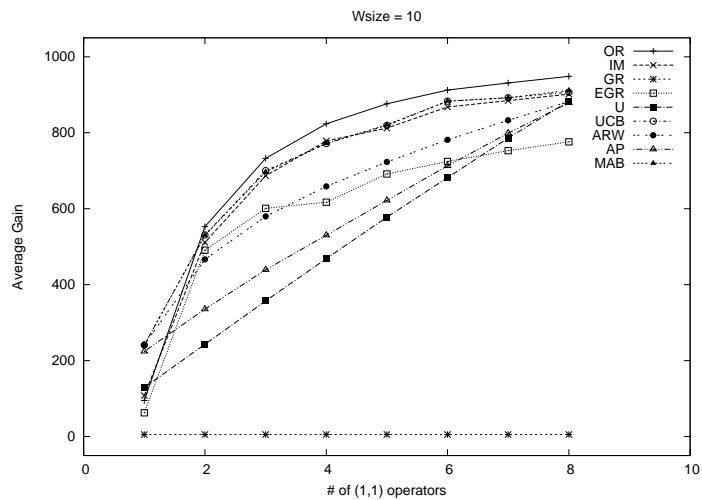}
\end{center}
\caption{Experiments with $wsize=10$}
\label{graph:win10}
\end{figure}

\begin{itemize}
\item
Considering the first column of tables \ref{res1} and \ref{res2}, i.e. scenario with $N_{op} = 1$, it should be noted that the suboptimal optimization solution presented in Section \ref{sec:nsbs} provides better results than $OR$. This is due to the fact that $OR$ is a myopic oracle while the suboptimal circular policy based on circular scenarios takes into account sequences of applications (in a dynamic programming fashion).  
\item
As expected, greedy strategies cannot insure a good schedule of the different operators in this non stationary context. 
\item
$UCB$ and $MAB$ are equivalent since no restart of the learning process is used here --- all $MAB$ parameters are set to $0$. Note that these parameters are useful when the distribution of gains is varying according to epochs (epochs based scenarios previously described). Here, since gains change continuously, such restart strategy is not efficient --- or would induce too much noise in the learning process. This has been checked by experiments. Note that here the standard UCB can be used directly without any reward normalization factor, since all rewards range from $0$ to $1$.   
\item 
According to the results mentioned in Section \ref{sec:nsbs}, the uniform choice $U$ provides the same results independently from the size of the window. 
\item 
ARW and AP obtain comparable results, with a slight superiority of ARW. Both policies give more importance to the best operator along the search. Such a strategy is not necessary well suited for this problem, but is rather efficient when $N_1$ is low. The balance adjustment between $\mathds{O}$ and $\mathds{1}$ operators is insured by the $p_{\min}$ parameter. Note that these policies obtain similar results independently from the size of the window, which seems to mean that their behaviour is close to a uniform choice, but restricted to the most efficient operators. When $N_1$ increases, they become indeed equivalent to U. We have checked that, studying the sequence of selected operators, no clear repeated sequence of operators can be observed ; however the total amount of $\mathds{1}$ operators used remains similar in the different problem configurations. 
\item
Concerning UCB, MAB and IM, we may distinguish several different ranges of results:
\begin{itemize}
\item
When $N_1=1$, UCB (and MAB, but in the following, we will only mention UCB) and IM cannot insure a good management of the balance between $\mathds{O}$ and $\mathds{1}$ operators, certainly due to their exploration component --- ie. noise in IM and right member in UCB formula (see section \ref{sec:polucb}). 
. 
\item
When $1 \leqslant N_1 \leqslant wsize$, UCB seems to provide better results than IM on most instances. While $N_1$ increases, the gap between the policies reduces. Moreover, the policies are more and more close to the oracle. 
\item 
When $N_1 = wsize+1$, one observes that OR is able to compute an optimal schedule (1000). Indeed, it is possible here to alternate between  $\mathds{1}$ operators only. We also observe that IM is also able to increase its performance, and becomes then better than UCB. 
\item
When $N_1 > wsize+1$, the problem becomes easier since a uniform choice U is then a reasonably efficient policy. Here, IM and UCB have a performance close to OR. 
\end{itemize}
\item
We may remark that the behaviour of the different policies is almost the same when dimensions of the problem increase, shifted according to the values of $N_1$ and $wsize$. We have conducted experiments on other dimensions (see Figure \ref{graph:win10}) with similar observations. 

\end{itemize}

\section{Conclusion}

In this paper, we have proposed a new model for simulating non stationary operators in search algorithms that should alternate between intensification and diversification stages in their search processes. The abstract model that is defined here may serve to evaluate the performance of operator selection policies in these search algorithms. We proposed here an experimental studies of different classic operator selection policies in order to highlight their respective advantages and drawbacks in such search scenarios. 

Our model can be considered as possible a surrogate model in order to design new adaptive search algorithms that aim at soving general optimization problems without focusing on specific dedicated operators or search heuristics. For a reinforcement learning point of view, our model corresponds to specific restless bandit problems that could be used to model different real applications as soon as the efficiency of a given action decreases according to successive frequent uses.

\bibliographystyle{elsarticle-harv}
\bibliography{bm}
\end{document}